\pdfoutput=1

\documentclass[11pt]{article}

\usepackage{acl}

\usepackage{times}
\usepackage{latexsym}

\usepackage[T1]{fontenc}

\usepackage[utf8]{inputenc}

\usepackage[letterspace=-80]{microtype}

\usepackage{bm}
\usepackage{graphicx}
\usepackage{multirow}
\usepackage{amsmath}

\usepackage{tabularx}
\usepackage{booktabs}

\usepackage{tikz}
\usepackage{xcolor}

\usepackage{hyperref}

\title{CoDA21:
Evaluating Language Understanding Capabilities
of NLP Models
With
Context-Definition Alignment}

\author{Lütfi Kerem Senel, Timo Schick {\normalfont and} Hinrich Schütze\\
   Center for Information and Language Processing (CIS), LMU Munich, Germany \\
   \texttt{lksenel@gmail.com}, \texttt{schickt@cis.lmu.de}
}

\newcommand\thename{CoDA21}
\newcommand\wngroup{\thename{} group}
\newcommand\placeholder{\textsf{<xxx>}}

\definecolor{c0}{RGB}{31,78,121} 
\usetikzlibrary{positioning,arrows,arrows.meta}

\newcounter{notecounter}
\newcommand{\enotesoff}{\long\gdef\enote##1##2{}}
\newcommand{\enoteson}{\long\gdef\enote##1##2{{
\stepcounter{notecounter}
{\large\bf
\hspace{1cm}\arabic{notecounter} $<<<$ ##1: ##2
$>>>$\hspace{1cm}}}}}

\enoteson
\enotesoff

\def\uprm#1{\mbox{$^{\hbox{\scriptsize #1}}$}}
\long\def\eat#1{\ignorespaces}

\begin{document}
\maketitle
\begin{abstract}

Pretrained language models (PLMs) have achieved superhuman
performance on many benchmarks, creating a need for harder
tasks.
We
introduce \thename{} (Context
Definition Alignment), a challenging benchmark that measures
natural language understanding (NLU) capabilities of PLMs:
Given a definition and a context each for $k$ words, but not
the words themselves, the task is to align the $k$
definitions with the $k$ contexts. \thename{} requires a
deep understanding of contexts and definitions, including
complex inference and world knowledge.  We find that there
is a large gap between human and PLM performance, suggesting
that \thename{} measures an aspect of NLU that is not
sufficiently covered in existing benchmarks.\footnote{\textcolor{black}{Our dataset and code are available at \url{https://github.com/lksenel/CoDA21}}}

\end{abstract}

\section{Introduction}

Increasing computational power along with the design and
development of large and sophisticated models that can take
advantage of enormous corpora has 
drastically advanced NLP.  For
many tasks,
finetuning pretrained transformer-based language
models \cite{vaswani17transformers, devlin-etal-2019-bert,
radford2018GPT} has
improved the state of the art considerably.
\enote{ts}{imo, the following sentence can be removed if we need space (I guess the purpose of pretraining is obvious to most readers)} 
Language models  acquire
knowledge during pretraining that is utilized during task-specific finetuning.
On benchmarks
that were introduced to encourage development of models
that do well on a diverse set
of NLU tasks
(e.g., GLUE
\cite{wang-etal-2018-glue}
and
SuperGLUE
\cite{wang2019superglue}),
these models now achieve
superhuman performance \cite{he2020deberta}.
\enote{hs}{I guess the following sentence is meant to
justify why we don't finetune? I very much doubt that the
reader will get this message.}
The pretrain-then-finetune approach usually requires a great
amount of labeled data, which is often not available or
expensive to obtain, and results in specialized models that
can perform well only on a single task.  Recently, it was
shown that generative language models can be applied to many
tasks without finetuning when the task is
formulated as text generation and the PLM is queried with a
natural language prompt  \cite{Radford19GPT2,brown20GPT3}.

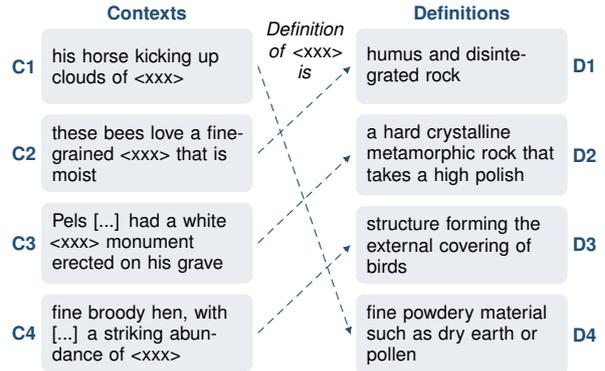
\begin{figure}
    \centering
    \tikzset{
    	every node/.style={
    		outer sep=0, font=\sffamily\scriptsize
    	},
    	pattern/.style={
    		text height=1.5ex, outer sep=0, inner sep=0, text width=1.22cm, align=center,
    	},
    	context/.style={
    		minimum height=1cm, text width=2.5cm, outer sep=2pt, inner sep=4pt, rounded corners=3pt, fill=c0!10, thick
    	},
        definition/.style={
    		minimum height=1cm, text width=2.5cm, outer sep=2pt, inner sep=4pt, rounded corners=3pt, fill=c0!10, thick
    	},
        context-label/.style={
    		text=c0, inner sep=0, outer sep=0
	    },
	    definition-label/.style={
	    	text=c0, inner sep=0, outer sep=0
	    },
    	arrow/.style={
    		->,>={Latex[length=1mm, width=1mm]}, draw=c0, densely dashed
    	},
    }
    \begin{tikzpicture}
    \node[context](c1){his horse kicking up clouds of \placeholder{}};
    \node[context, below=0cm of c1](c2){these bees love a fine-grained \placeholder{} that is moist};
    \node[context, below=0cm of c2](c3){Pels [...] had a white \placeholder{} monument erected on his grave};
    \node[context, below=0cm of c3](c4){fine broody hen, with [...] a striking abundance of \placeholder{}};
    
    \node[context-label, left=0cm of c1]{\textbf{C1}};
    \node[context-label, left=0cm of c2]{\textbf{C2}};
    \node[context-label, left=0cm of c3]{\textbf{C3}};
    \node[context-label, left=0cm of c4]{\textbf{C4}};
    
    \node[pattern, right=0cm of c1, yshift=0.25cm](pattern){\emph{Definition of} \placeholder{} \emph{is}};
   
    \node[definition, right=1.22cm of c1](d1){humus and disintegrated rock};
    \node[definition, below=0cm of d1](d2){a hard crystalline metamorphic rock that takes a high polish};
    \node[definition, below=0cm of d2](d3){structure forming the external covering of birds};
    \node[definition, below=0cm of d3](d4){fine powdery material such as dry earth or pollen};
    
    \node[definition-label, right=0cm of d1]{\textbf{D1}};
    \node[definition-label, right=0cm of d2]{\textbf{D2}};
    \node[definition-label, right=0cm of d3]{\textbf{D3}};
    \node[definition-label, right=0cm of d4]{\textbf{D4}};
    
    \node[context-label, above=0.05cm of c1](contexts-headline){\textbf{Contexts}};
    \node[definition-label, above=0.05cm of d1](definitions-headline){\textbf{Definitions}};
    
    \path[] (c1.east) edge[arrow] (d4.west);
    \path[] (c2.east) edge[arrow] (d1.west);
    \path[] (c3.east) edge[arrow] (d2.west);
    \path[] (c4.east) edge[arrow] (d3.west);
    \end{tikzpicture}

    \caption{The \thename{} task is to find the correct alignment between contexts and definitions:
    \textbf{C1}-\textbf{D4}, 
    \textbf{C2}-\textbf{D1}, 
    \textbf{C3}-\textbf{D2}, 
    \textbf{C4}-\textbf{D3}. 
     The target words (\textbf{C1}:``dust'', \textbf{C2}:``soil'',
    \textbf{C3}:``marble'', \textbf{C4}:``feathers''; not
    provided to the model) are replaced with a placeholder \placeholder{}.}
    \label{fig:CoDA_task}
\end{figure}

Motivated by recent progress in zero-shot learning with generative models as well as the need for more challenging benchmarks 
that test language understanding of language models, we introduce \thename{} (\textbf{Co}ntext \textbf{D}efinition \textbf{A}lignment), a
difficult benchmark that measures NLU capabilities of PLMs 
\textcolor{black}{for the English language}.
Given a definition and a context each for $k$ words, but not the words themselves,
the task is to align the $k$ definitions with the $k$ contexts.  
In other words, for each definition, the context in which the defined word is most likely to occur has to be identified.
This requires (i) understanding the definitions,  (ii) understanding the contexts, and (iii) the ability to match the two.  
Since the target words are not given, a model must be able to distinguish subtle meaning differences between different contexts/definitions to be successful.  
To illustrate the difficulty of the task, Figure~\ref{fig:CoDA_task} shows a partial example for
$k=4$
(see Table \ref{tab:sample} in the supplementary for the full example).
We see that both complex inference (e.g., \texttt{<XXX>} can give rise to a cloud by being kicked up
$\Rightarrow$ \texttt{<XXX>} must be dry
$\Rightarrow$ \texttt{<XXX>} can be dust, but not soil) and
world knowledge (what materials are typical for monuments?) are required for \thename{}. 

We formulate the alignment task as a text prediction task
and evaluate, without finetuning, three PLMs on \thename{}:
BERT \cite{devlin-etal-2019-bert}, RoBERTa \cite{liu19RoBERTa} and
GPT-2 \cite{Radford19GPT2}. 
Poor performance of the PLMs
and a large gap between human and PLM
performance suggest that \thename{} is an important
benchmark for designing models with better NLU capabilities.

\section{\thename{}}


\subsection{Dataset}
We construct \thename{} by first deriving a set ${\cal G}$ of \emph{synset
  groups} $\{G_1,G_2,\ldots\}$ from Wordnet \cite{miller95wordnet}. A synset
group $G_i$ is
a group of synsets whose meanings are close enough to be
difficult to distinguish (making the task hard), 
but not so close that they become indistinguishable for
human and machine. In a second step, each synset group $G_i$
is converted into a \emph{\wngroup} $G_i^+$ --  a set of triples,
each consisting of the synset, its definition, and a corpus
context. A \wngroup\ can be directly used for one instance
of the \thename{} task.

\textbf{Synset groups.}
Each synset group $G$ consists of $5 \leq k \leq 10$
synsets.
To create a synset group, we start with a \emph{parent synset} 
$\hat{s}$ and construct a co-hyponym group $\bar{G}(\hat{s})$ of its children:
\begin{equation*}
    \bar{G}(\hat{s}) =
    \{s \mid  s < \hat{s}, s\notin{ D} \}
\end{equation*}
where $<$ is the hyponymy relation between synsets and ${
D}$ is the set of synsets that have already been added to a
synset group.
The intuition for grouping synsets with a common parent is that words sharing a hypernym are difficult
to distinguish (as opposed to randomly selected
words). 

We iterate $\hat{s}$ through all nouns and verbs in WordNet.
At each iteration, we get all hyponyms of $\hat{s}$ that
have not been previously added to a synset group; not reusing
a synset ensures that different \thename{} subtasks are not
related and so no such relationships can be exploited.

\textcolor{black}{
We extract synset groups from co-hyponym groups by splitting 
them into multiple chunks of size $k$.
In an initial exploration, we found that the task is hard to
solve for human subjects if two closely related hyponyms are
included, e.g., ``clementine'' and ``tangerine''. We
therefore employ clustering to assemble a set of mutually
dissimilar hyponyms. We first compute a
sentence embedding for each hyponym definition using
the \emph{stsb-distilbert-base} Sentence Transformer model\footnote{\url{https://huggingface.co/sentence-transformers/stsb-distilbert-base}}.
We then 
cluster the embeddings using complete-link
clustering, combining the two most dissimilar clusters in
each step.
We stop merging before the biggest cluster exceeds the
maximum group size ($k=10$) or before the similarity between the
last two combined clusters exceeds the maximum similarity
($\theta = 0.8$).
The largest cluster $G$ is added to the set $\cal G$
of synset groups.
We then iterate the steps of
(i) removing the synsets in the previous largest cluster ${G}$ from
$\bar{G}(\hat{s})$ and (ii) running complete-link clustering
and adding the resulting largest cluster $G$
to $\cal G$
until
fewer than five synsets remain in
${\bar G}(\hat{s})$ or no cluster can be formed
whose members have a 
similarity of less than $\theta$.}

\textbf{\wngroup s.}
For each synset $s$, we extract its definition
$d(s)$ from WordNet and a context $c(s)$ in which it occurs from
SemCor\footnote{We do not consider synsets without contexts in SemCor.} \citep{miller1994semcor}.  SemCor
is an
English corpus tagged with WordNet senses.
Let $C(s)$ be the set of contexts of
$s$ in SemCor.
If $|{C}(s)| > 1$,
we use as $c(s)$ the context in which
\emph{bert-base-uncased} \textcolor{black}{predicts
$\textbf{s}$ with the
highest log probability when it is masked, where $\textbf{s}$ is the word tagged with the sense $s$}\footnote{\textcolor{black}{We average the probabilities when $\textbf{s}$ is tokenized to multiple tokens.}} -- this favors contexts that
are specific to the meaning of the synset.
Finally, we convert each synset group $G_i$ in $\cal G$ to a 
\wngroup{} $G_i^+$:
\[
G_i^+ = \{ (s_j, d(s_j),
c(s_j)) \mid s_j \in G_i \}
\]
That is, a \wngroup{} $G_i^+$ is a set
of triples of
sense, definition and context.
In
PLM evaluation, each \wngroup{} ${G_i^+}$ gives rise to one
context-definition alignment subtask.

We name the resulting
dataset \textit{\thename{}-noisy-hard}: \textit{noisy}
because if $|{C}(s)|$ is small, the selected context
may not be informative enough to identify the matching definition;
\textit{hard}
because the synsets in a \wngroup{} are taxonomic sisters,
generally with
similar meanings despite the clustering-based limit 
on definition similarity.  We construct a \textit{clean}
version of the dataset by only using synsets with $|{C}(s)| \geq 5$.  We also construct an \textit{easy} version
by taking the  ``hyponym grandchildren'' $s$ of a parent
synset $\hat{s}$ ($s<l \wedge l<\hat{s}$) instead of its
hyponym children.  This reduces the similarity of  synsets in a \wngroup{},
making the task easier.  
Table \ref{tab:coda_counts} gives dataset statistics.

\begin{table}[]
    \centering
    \small
    \begin{tabular}{lrrrr}
    \toprule
       \multirow{2}{*}{\textbf{Dataset}} & \multicolumn{2}{c}{noun} & \multicolumn{2}{c}{verb} \\
        & \# of ${G}$ & USC  & \# of ${G}$ & USC  \\
       \midrule
        \thename{}-\textit{clean-hard} & 106 & 740 & 102 & 711 \\
        \thename{}-\textit{clean-easy} & 274 & 1999 & 103 & 758 \\
        \thename{}-\textit{noisy-hard} & 691 & 4633 & 350 & 2527 \\
        \thename{}-\textit{noisy-easy} & 1188 & 8910  & 370 & 2766 \\
        \bottomrule
    \end{tabular}
    \caption{\textcolor{black}{\wngroup{} (${G}$) statistics, USC: Unique Synset Count}}
    \label{tab:coda_counts}
\end{table}



\subsection{Alignment}
Recall the \thename{} task: given a definition and a context
each for $k$ words (but not the words themselves), align the
$k$ definitions with the $k$ contexts. That is, we are
looking for a bijective function (a one-to-one
correspondence) between definitions and contexts.
Our motivation in designing the task is that we want a hard
task (which can guide us in developing
stronger natural language understanding 
models), but also a task that is solvable by humans. Our
experience is that humans can at least partially solve the
task by finding a few initial ``easy'' context-definition
matches, removing them from the definition/context sets and
then match the smaller remaining number of
definitions/contexts.


The number of context-definition pairs scales quadratically ($O(k^2)$)
with $k$ and the number of alignments factorially
($O(k!)$). We restrict $k$ to $k \leq 10$ to make sure that
we do not run into computational problems and that humans do
not find the task too difficult.

\textcolor{black}{
In order to connect contexts to definitions without using the target words, we replace the target words by a made-up word. 
This setup resembles the incidental vocabulary acquisition process in humans.
}
Let $t$ be a target word, $c$ a context in which $t$ occurs and $m$ a made-up word. To test PLMs on \thename{}, we use the following pattern\footnote{\textcolor{black}{When the target word is a verb (i.e., verb subset of a \thename{} dataset), we add ``to'' at the end of our pattern.
}}:
\begin{align*}
Q(c, m) & = c_m\text{ Definition of $m$ is}
\end{align*}
where $c_m$ is $c$ with 
occurrences of 
$t$ replaced by $m$. 

We calculate the \emph{match score} of a
context-definition pair $(c,d)$ as $\log P(d \mid Q(c,m))$, i.e., log generation probability of
the definition $d$ conditioned on $Q(c,m)$. 
Our objective is to maximize the sum of the $k$ match scores in
an alignment. We find the best alignment by exhaustive search.
Accuracy for a
\wngroup{} $G_i^+$ is then the accuracy of its best alignment, i.e., the number of contexts in $G_i^+$ that are aligned with the correct definition, divided by the total number of contexts $|G_i^+|$.

\subsection{Baselines}
We calculate  $ P(d \mid Q(c,m))$ for 
a masked language model (MLM) $M$ and an autoregressive language model (ALM) $A$ as follows:
\begin{align*}
    P_M(d \mid Q') & = \mbox{$\prod_{i=1}^{|d|}P(d_i \mid Q',d_{-i})$}\\
    P_A(d \mid Q') & = \mbox{$\prod_{i=1}^{|d|}P(d_i \mid Q',d_1,\ldots,d_{i-1})$}
\end{align*}
where $Q'= Q(c,m)$, 
$d_i$ is the $i\uprm{th}$ word in definition $d$ and
$d_{-i}$ is the definition with the $i\uprm{th}$ word masked.

We evaluate the  MLMs BERT and RoBERTa and
the ALM GPT-2. We experiment with both
base and large versions of BERT and RoBERTa and with all
four sizes of GPT-2 (small, medium, large, xl), for a total of
eight models, to investigate the effect of model size on performance. 

The made-up word $m$ should ideally be unknown so that it does 
not bias the PLM in any way.
However, there are no truly unknown words for the models we
investigate due to the word-piece tokenization they apply to
the input. Any made-up word that is completely meaningless
to humans will have a representation in the models' input
space based on its tokenization. 
To minimize the risk that the meaning of the made-up word
may bias the model,
we use $m = $ \textit{bkatuhla}, a word with an empty search
result on Google that
most likely never appeared in the models' pretraining corpora.

In addition to PLMs, we also evaluate 2 recent sentence transformer 
models\footnote{\url{https://www.sbert.net/docs/pretrained_models.html}} \cite{reimers-gurevych-2019-sentence},  
\emph{paraphrase-mpnet-base-v2} (mpnet) and \emph{paraphrase-MiniLM-L6-v2} (MiniLM), 
and fastText static embeddings\footnote{We use the \emph{crawl-300d-2M-subword} 
model from \lsstyle\url{https://fasttext.cc/docs/en/english-vectors.html}} \cite{mikolov-etal-2018-advances}.
To calculate the match score of a context-definition pair, we first 
remove the target word from the context and represent contexts and definitions 
as vectors. For sentence transformers, we obtain these vectors by simply encoding the input sentences. 
For fastText, we average the vectors of the words in contexts and definitions.
We then calculate the match score as the cosine similarity of context and definition vectors. 

\section{Results}

Table~\ref{tab:results} presents average accuracy of the investigated 
models on the four \thename{} datasets. 
As can be seen, fastText performs only slightly better than random. 
MLMs also perform better than random chance by only a small margin.  
This poor performance can be partly explained by the generation style 
setup we use, which is not well suited for masked language models. 
Even the smallest
GPT-2 model performs considerably better than RoBERTa-large, the best performing MLM.
Performance generally improves with  model size. 
GPT-2$_{xl}$ achieves the best results among the LMs 
on almost all datasets. Interestingly, sentence transformer 
\emph{all-mpnet-base-v2} performs comparably to GPT-2$_{xl}$ on most datasets 
despite its simple, similarity based matching compared to generation based matching
of GPT-2 models. 
\emph{Based on this observation it can be argued that current state-of-the-art 
language models fail to perform complex, multi-step reasoning and inference 
which are necessary to solve the \thename{} tasks.}
Overall, MLMs perform slightly better on verbs than nouns while the 
converse is true for  GPT-2.
As expected, all models perform better on the
\emph{easy} datasets.  
Performance on \emph{noisy} and \emph{clean} datasets are comparable; 
this indicates that our contexts are of high quality even for the synsets 
with only a few contexts.

\def\colspace{0.075cm}
\def\doublecolspace{0.325cm}
\def\mrspace{0.325cm}

\begin{table}
	\small
	\centering
	\begin{tabular}{l@{\hspace{\doublecolspace}}c@{\hspace{\colspace}}c@{\hspace{\doublecolspace}}c@{\hspace{\colspace}}c@{\hspace{\doublecolspace}}c@{\hspace{\colspace}}c@{\hspace{\doublecolspace}}c@{\hspace{\colspace}}c@{\hspace{\doublecolspace}}c}
		\toprule
		\multirow{2}{*}{{}} 
		& \multicolumn{2}{c@{\hspace{\mrspace}}}{{clean}} 
		& \multicolumn{2}{c@{\hspace{\mrspace}}}{{clean}} 
		& \multicolumn{2}{c@{\hspace{\mrspace}}}{{noisy}} 
		& \multicolumn{2}{c@{\hspace{\mrspace}}}{{noisy}}  
		& \multirow{2}{*}{{S20}}\\
		\multirow{2}{*}{{}} 
		& \multicolumn{2}{c@{\hspace{\mrspace}}}{{hard}}
		& \multicolumn{2}{c@{\hspace{\mrspace}}}{{easy}}
		& \multicolumn{2}{c@{\hspace{\mrspace}}}{{hard}}
		& \multicolumn{2}{c@{\hspace{\mrspace}}}{{easy}}  &
		\\
		\midrule
		\textbf{Model} & {N} & {V} & {N} & {V} & {N} & {V} & {N} & {V} & {N} \\
		\midrule
		BERT$_b$ & .20 & .21 & .22 & .25 & .21 & .22 & .22 & .24  & .24 \\ 
		BERT$_l$ & .22 & .22 & .19 & .21 & .19 & .20 & .20 & .20  & .22 \\ 
		RoBERTa$_b$ & .24 & .26 & .26 & .32 & .25 & .25 & .28 & .27  & .29 \\ 
		RoBERTa$_l$ & .26 & .30 & .30 & .30 & .27 & .29 & .30 & .33  & .29 \\
		GPT-2$_s$ & .31 & .32 & .42 & .40 & .35 & .32 & .40 & .36  & .35 \\ 
		GPT-2$_m$ & .37 & .35 & .45 & .39 & .38 & .35 & .43 & .39  & .39 \\ 
		GPT-2$_l$ & .38 & .34 & .47 & \textbf{.42} & .39 & \textbf{.37} & \textbf{.46} & .41  & .47 \\ 
		GPT-2$_{xl}$ & \textbf{.42} & .36 & \textbf{.49} & \textbf{.42} & \textbf{.40} & .36 & \textbf{.46} & \textbf{.43}  & .48 \\
		\midrule
        mpnet & \textbf{.42} & \textbf{.39} & .48 & \textbf{.42} & \textbf{.40} & \textbf{.37} & \textbf{.46} & .40 & .51 \\
        MiniLM & .35 & .34 & .40 & .36 & .34 & .30 & .38 & .32 & .34 \\
        fastText & .18 & .17 & .20 & .20 & .18 & .18 & .18 & .18 & .17 \\
		Random & .15 & .15 & .14 & .14 & .16 & .15 & .14 & .14 & .14 \\
		\midrule
		Human & -- & -- & -- & -- & -- & -- & -- & -- & \textbf{.86} \\
		\bottomrule 
	\end{tabular}
	\caption{Average accuracy on the noun (N) and verb (V) subsets of \thename{} for eight PLMs,
		two sentence transformers, fastText embeddings
		and (on S20) for humans}
	\label{tab:results}
\end{table}

\textbf{Human performance on \thename{}.}
We asked two NLP PhD students\footnote{Both
are proficient (though not native) English speakers.}
to solve the task on S20, a random sample of size
20 from the noun part
of \thename{}-\textit{clean-easy}.  Table \ref{tab:results}
shows results on S20 for these two subjects and our models.
Human performance
is 0.86 -- compared to 0.48
for GPT-2$_{xl}$, the best performing model.  This
difference indicates that there is a large gap in
NLU competence between current language
models and humans and that \thename{} is a good benchmark to
track progress on closing that gap.

\textcolor{black}{
To investigate the \textbf{effect of the made-up word} $m$, we
experiment with several other words 
on the noun part of \thename{}-\emph{clean-easy}. 
Specifically, we investigate another nonce word ``opyatzel'', a single letter ``x'' and two frequent words ``orange'' and ``cloud''.
Table \ref{tab:effect_of_made_up} shows the results of the models for different made-up words.
MLMs do not show significant variability in performance, and perform comparably poor for all words tried. 
GPT2 versions, which perform considerably better than MLMs on \thename{}, perform similarly for the two nonce words and single letter ``x'', which do not have a strong meaning.
Their performance drops significantly when the 
two frequent words are used as the made-up word, due to the effect of prior knowledge models have about these words.
}

\begin{table}
	\small
	\centering
	\begin{tabular}{l@{\hspace{\doublecolspace}}c@{\hspace{\doublecolspace}}c@{\hspace{\doublecolspace}}c@{\hspace{\doublecolspace}}c@{\hspace{\doublecolspace}}c}
		\toprule
        \textbf{Model}
		& bkatuhla 
        & opyatzel
        & x
        & cloud
        & orange
		\\
		\midrule
		BERT$_b$ & .22 & .22 & .22 & .23 & .22 \\ 
		BERT$_l$ & .19 & .19 & .20 & .20 & .19 \\ 
		RoBERTa$_b$ & .26 & .27 & .26 & .28 & .28 \\ 
		RoBERTa$_l$ & .30 & .30 & .29 & .30 & .29 \\
		GPT-2$_s$ & .42 & .43 & .41 & .39 & .39 \\ 
		GPT-2$_m$ & .45 & .42 & .43 & .40 & .41 \\ 
		GPT-2$_l$ & .47 & .46 & .47 & .41 & .42 \\ 
		GPT-2$_{xl}$ & .49 & .44 &.45 & .40 & .41  \\
		\bottomrule 
	\end{tabular}
	\caption{\textcolor{black}{Average accuracy of eight PLMs on the noun subsets of \thename{}\textit{-clean-easy} using various words as the made-up word.}}
	\label{tab:effect_of_made_up}
\end{table}

\textcolor{black}{
To investigate the \textbf{effect of the pattern}, we compared our pattern $Q(c,m)$ with two 
alternative patterns by evaluating GPT-2$_{xl}$ 
on the noun part of \thename{}-\emph{clean-easy}.
Patterns and the evaluation results are shown in 
Table \ref{tab:pattern}. The results suggest that 
the effect of the pattern on performance is minimal.
}

\begin{table}[h]
    \centering
    \small
    \begin{tabular}{lc}
    \toprule
        \textbf{Pattern} & \textbf{Acc} \\
        \midrule
        $c_m$ Definition of $m$ is & 0.49 \\
        $c_m$ $m$ is defined as & 0.51 \\
        $c_m$ $m$ is & 0.49 \\
        \bottomrule
    \end{tabular}
    \caption{Effect of the pattern on the performance of GPT2-$_{xl}$ on the noun part of \thename{}-\textit{clean-easy}}
    \label{tab:pattern}
\end{table}

\textcolor{black}{
\textbf{Effect of the alignment setup.}
We constructed \thename{} as an alignment dataset which uses the fact that 
matching between the definitions and contexts is one-to-one. 
This setup makes the task more intuitive and manageable for humans. 
However, context-definition match scores can be used to evaluate 
models on \thename{} samples also without the alignment setup 
by simply picking context-definition pairs with the highest match score 
for each definition. We additionally evaluated GPT-2$_{xl}$ model on 
\thename{}-\textit{clean-easy} dataset using this simple matching approach which yielded 
0.38 average accuracy compared to the 0.49 accuracy achieved with the alignment setup.
This result suggests that language models can also make use of the alignment style 
evaluation, similar to humans.
}

\begin{figure}[t]
    \centering
    \includegraphics[width=\columnwidth]{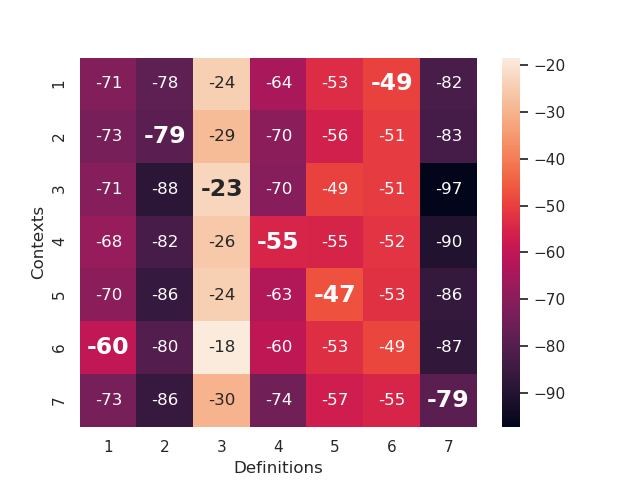}
    \caption{Match scores from GPT2-xl model for the context definition pairs for the sample given in Table \ref{tab:sample}.
    Match scores shown in bold correspond the context-definition pairs that are in the predicted alignment by the model that yields maximum total match score.}
    \label{fig:heat_map}
\end{figure}

\textcolor{black}{
Table \ref{tab:sample} (in the Appendix) presents a sample of size 7 from the noun part of the 
\thename{}-\textit{clean-easy} dataset.
Figure \ref{fig:heat_map} displays all 49 match scores of the context-definition 
pairs for this sample obtained using GPT-2$_{xl}$. 
5 of the 7 definitions (2,3,4,5,7) are matched with correct contexts with the alignment setup
while 4 definitions (4,5,6,7) are matched correctly for the simple matching setup.
Alignment setup enabled the model to match second and third definitions with their 
corresponding contexts even though their match scores are not the highest ones. 
}

\textcolor{black}{
To get a better sense of why the task is hard for PLMs, we
give an example, from the \thename{} subtask in 
Figure~\ref{fig:CoDA_task} (also Figure \ref{fig:heat_map} and Table \ref{tab:sample} refer to the same subtask), 
of a context-definition match that is
scored highly by GPT-2$_{xl}$, but is not correct. \textbf{Context:} 
``these bees love a fine-grained \texttt{<XXX>} that is moist''. \textbf{Definition:} 
``fine powdery material such as dry earth or pollen''. (context 6 and definition 1 in Figure \ref{fig:heat_map})
GPT-2$_{xl}$ most likely gives a high score because it has learned that
\textit{bees} and \textit{pollen} are associated. It does
not understand that the mutual exclusivity of
``moist'' and ``powdery'' makes this a bad match.
}

\enote{hs}{i agree: we should have some qualitative analysis

the easiest thing to do is probably your second suggestion:
find some really low match scores for correct
context-definition pairs and then come up with some
qualitative explanation for why the scores are low}

\section{Related Work}
There are
many datasets \cite{levesque2012WSC, rajpurkar-etal-2016-squad, williams-etal-2018-broad} for evaluating language understanding of models.
Many adopt a text prediction setup:
Lambada \cite{paperno-etal-2016-lambada} evaluates the
understanding of discourse context,
StoryCloze \cite{mostafazadeh-etal-2016-corpus} 
evaluates commonsense knowledge and so does
HellaSwag \cite{zellers-etal-2019-hellaswag}, but 
examples were adversarially mined. 
LAMA \cite{petroni-etal-2019-language} tests the factual knowledge
contained in PLMs.
In contrast to this prior work, \thename{} goes beyond
prediction by requiring the matching of pieces of text.
WIC \cite{pilehvar-camacho-collados-2019-wic} is also based on matching, but
\thename{} is more complex (multiple contexts/definitions as
opposed to a single binary match decision) and is not
restricted to ambiguous words.
WNLaMPro \cite{Schick20rareWords} evaluates knowledge of
subordinate relationships between words,
and WDLaMPro \cite{senel-schutze-2021-wink} understanding of words
using dictionary definitions. Again, matching multiple
pieces of text with each other is much harder and therefore
a promising task for benchmarking NLU.

\enote{hs}{potentially reinstate material from here:

\thename{} is most similar to
which tests word understanding on dictionary definitions.
The \thename{} alignment combines
the two individual matching tasks proposed
in \cite{senel-schutze-2021-wink}
by using the fact that the
matching between the contexts and definitions is one to one.
}

\section{Conclusion}
We introduced \thename{}, a new challenging benchmark that
tests natural language understanding capabilities of PLMs.
Performing well on \thename{} requires detailed
understanding of contexts, performing complex inference and
having world knowledge, which are crucial skills for
NLP.  All models we
investigated perform clearly worse than humans, indicating
a lack of these skills in the current state of the art in NLP. \thename{}
therefore is a promising benchmark for guiding  the
development of models with stronger NLU competence.

\bibliography{anthology,CoDA}
\bibliographystyle{acl_natbib}

\clearpage

\appendix

\section{Appendices}

\subsection{\wngroup{} examples}

\begin{table*}
    \centering
    \begin{tabularx}{\textwidth}{|l|X|}
        \hline
        \textbf{Hidden word} & \textbf{Context} \\ \hline
        dust & 1. He came spurring and whooping down the road , his horse kicking up clouds of <XXX> , shouting : \\
        marble & 2. Pels also sent a check for \$ 100 to Russell 's widow and had a white <XXX> monument erected on his grave .\\
        wastewater & 3. The high cost of land and a few operational problems resulting from excessive loadings have created the need for a <XXX> treatment system with the operational characteristics of the oxidation pond but with the ability to treat more organic matter per unit volume . \\
        feathers & 4. It was a fine broody hen , white , with a maternal eye and a striking abundance of <XXX> in the under region of the abdomen . \\
        fraction & 5. It was then distilled at least three times from a trap at - 78 ` to a liquid air trap with only a small middle <XXX> being retained in each distillation . \\
        soil & 6. The thing is that these bees love a fine-grained <XXX> that is moist ; yet the water in the ground should not be stagnant either . \\
        cards & 7. And the coffee shop on Drexel Street , where the men spent their evenings and Sundays playing <XXX> , had a rose hedge beneath its window . \\ \hline
        \textbf{Synset} & \textbf{Definition} \\ \hline
        dust.n.01 & 1. fine powdery material such as dry earth or pollen that can be blown about in the air \\
        marble.n.01 & 2. a hard crystalline metamorphic rock that takes a high polish; used for sculpture and as building material\\
        effluent.n.01 & 3. water mixed with waste matter \\
        feather.n.01 & 4. the light horny waterproof structure forming the external covering of birds \\
        fraction.n.01 & 5. a component of a mixture that has been separated by a fractional process \\
        soil.n.02 & 6. the part of the earth's surface consisting of humus and disintegrated rock \\
        card.n.01 & 7. one of a set of small pieces of stiff paper marked in various ways and used for playing games or for telling fortunes \\ \hline
    \end{tabularx}
    \caption{A sample \thename{} question taken from the noun part of the \thename{}-\textit{clean-easy} dataset.
    The synsets are grandchildren of the parent synset `material.n.01' whose definition is ``the tangible substance that goes into the makeup of a physical object''.}
    \label{tab:sample}
\end{table*}

\begin{table*}
    \centering
    \begin{tabularx}{\textwidth}{|l|X|}
        \hline
        \textbf{Hidden word} & \textbf{Context} \\ \hline
        suggestion & 1. This was Madden 's <XXX> ; the police chief shook his head over it . \\
        concept & 2. The <XXX> of apparent black-body temperature is used to describe the radiation received from the moon and the planets .\\
        ideals & 3. Religion can summate , epitomize , relate , and conserve all the highest <XXX> and values - ethical , aesthetic , and religious - of man formed in his culture . \\
        reaction & 4. That much of what he calls folklore is the result of beliefs carefully sown among the people with the conscious aim of producing a desired mass emotional <XXX> to a particular situation or set of situations is irrelevant . \\
        feeling & 5. He had an uneasy <XXX> about it . \\
        programs & 6. The Federal program of vocational education merely provides financial aid to encourage the establishment of vocational education <XXX> in public schools . \\
        meaning & 7. Indefinite reference also carries double <XXX> where an allusion to one person or thing seems to refer to another .\\ 
        theme & 8. Almost nothing is said of Charles ' spectacular victories , the central <XXX> being the heroic loyalty of the Swedish people to their idolized king in misfortune and defeat . \\ \hline
        \textbf{Synset} & \textbf{Definition} \\ \hline
        suggestion.n.01 & 1. an idea that is suggested \\
        concept.n.01 & 2. an abstract or general idea inferred or derived from specific instances \\
        ideal.n.01 & 3. the idea of something that is perfect; something that one hopes to attain \\
        reaction.n.02 & 4. an idea evoked by some experience \\
        impression.n.01 & 5. a vague idea in which some confidence is placed \\
        plan.n.01 & 6. a series of steps to be carried out or goals to be accomplished \\
        meaning.n.02 & 7. the idea that is intended \\ 
        theme.n.02 & 8. a unifying idea that is a recurrent element in literary or artistic work \\ \hline
    \end{tabularx}
    \caption{\textcolor{black}{A sample \thename{} question taken from the noun part of the \thename{}-\textit{clean-hard} dataset.
    The synsets are children of the parent synset `idea.n.01' whose definition is ``the content of cognition; the main thing you are thinking about''.}}
    \label{tab:hard_sample}
\end{table*}

\end{document}